\documentclass[conference]{IEEEtran}
\IEEEoverridecommandlockouts
\usepackage{cite}
\usepackage{amsmath,amssymb,amsfonts}
\usepackage{algorithmic}
\usepackage{graphicx}
\usepackage{textcomp}
\usepackage{xcolor,subfigure}
\usepackage{framed,multirow}
\usepackage{bm}
\usepackage{graphicx}
\usepackage{lettrine}
\usepackage{makecell}
\usepackage{url}
\usepackage{multicol}
\usepackage{rotating}

\def\etal{{\em et al.\/}\, }

\def\ie{\textit{i.e.}}
\def\BibTeX{{\rm B\kern-.05em{\sc i\kern-.025em b}\kern-.08em
    T\kern-.1667em\lower.7ex\hbox{E}\kern-.125emX}}
\begin{document}

\title{Coarse to Fine: Multi-label Image Classification with Global/Local Attention\\
\thanks{This work was supported by the Natural Science Foundation of China (Nos. 61472267, 61728205, 61502329, 61672371) and Primary Research \& Developement Plan of Jiangsu Province (No. BE2017663).}
}

\author{
%	\centeringt
%\IEEEauthorblockN{Fan Lyu}
%%\IEEEauthorblockA{\textit{School of Electronic \& Information Engineering}\\
%%\textit{Suzhou University of Science and Technology}\\
%%Suzhou, China \\
%%lvfan@post.ustu.edu.cn}
%\and
%\IEEEauthorblockN{Fuyuan Hu$^*$\thanks{$^*$Corresponding author}}
%%\IEEEauthorblockA{\textit{School of Electronic \& Information Engineering}\\
%%	\textit{Suzhou University of Science and Technology}\\
%%	Suzhou, China \\
%%	fuyuanhu@mail.ustu.edu.cn}
%\and
%\IEEEauthorblockN{Victor S. Sheng}
%%\IEEEauthorblockA{\textit{Computer Science Department}\\
%%\textit{University of Central Arkansas}\\
%%Arkansas, USA \\
%%ssheng@uca.edu}
%\and
%\IEEEauthorblockN{Zhengtian Wu}
%%\IEEEauthorblockA{\textit{School of Electronic \& Information Engineering}\\
%%\textit{Suzhou University of Science and Technology}\\
%%Suzhou, China \\
%%wzht8@mail.ustu.edu.cn}
%\and
%\IEEEauthorblockN{Qiming Fu}
%%\IEEEauthorblockA{\textit{Jiangsu Province Key Lab}\\
%%\textit{Intelligent Building Energy Efficiency}\\
%%Suzhou, China \\
%%fqm\_1@126.com}
%\and
%\IEEEauthorblockN{Baochuan Fu}
%%\IEEEauthorblockA{\textit{School of Electronic \& Information Engineering}\\
%%\textit{Suzhou University of Science and Technology}\\
%%Suzhou, China \\
%%fubc@mail.ustu.edu.cn}
Fan Lyu$^{1,2}$, Fuyuan Hu$^{2,*}$\thanks{$^*$Corresponding author}, Victor S. Sheng$^3$, Zhengtian Wu$^2$, Qiming Fu$^2$ and Baochuan Fu$^2$\\
~\\
\textsuperscript{\rm 1}\textit{Colledge of Intelligence and Computing, Tianjin University}\\
\textsuperscript{\rm 2}\textit{School of Electronic \& Information Engineering, Suzhou University of Science and Technology}\\
\textsuperscript{\rm 3}\textit{Department of Computer Science, Texas Tech University}\\
fanlyu@tju.edu.cn, \{fuyuanhu, wzht8, fubc\}@mail.usts.edu.cn, victor.sheng@ttu.edu
}
%\author{
%	%Afiliations
%	\textsuperscript{\rm 1}Colledge of Intelligence and Computing, Tianjin University\\
%	\textsuperscript{\rm 2}School of Electronic \& Information Engineering, Suzhou University of Science and Technology\\
%	\textsuperscript{\rm 3}Department of Computer Science and Engineering, University of South Carolina\\
%	%If you have multiple authors and multiple affiliations
%	% use superscripts in text and roman font to identify them.
%	%For example,
%	% Sunil Issar, \textsuperscript{\rm 2}
%	% J. Scott Penberthy, \textsuperscript{\rm 3}
%	% George Ferguson,\textsuperscript{\rm 4}
%	% Hans Guesgen, \textsuperscript{\rm 5}.
%	% Note that the comma should be placed BEFORE the superscript for optimum readability
%	%    2275 East Bayshore Road, Suite 160\\
%	%    Palo Alto, California 94303\\
%	% email address must be in roman text type, not monospace or sans serif
%	\{fanlyu, wangshuai201909, wfeng\}@tju.edu.cn, \{zihanye@post, fuyuanhu@mail\}.usts.edu.cn, songwang@cec.sc.edu
%	% See more examples next
%}

\maketitle

\begin{abstract}
In our daily life, the scenes around us are always with multiple labels especially in a smart city, i.e., recognizing the information of city operation to response and control.
%   Recognizing a complex multi-label image is very difficult due to several challenges.
%   First, in one image, multiple objects can be anywhere.
%   Then, regions of interest (ROI) in a multi-label image can be a very small part, and there exists much redundant information.
%   Last, the multiple labels of an image can have label dependencies among each other.
%   To address these issues, we mimic how human-beings observe the world and predict multiple labels for an multi-label image.
%---------------------------------
% 1. great progress
% 2. conventional drawacks
% 3. In this paper, we
% 4. several characteristic
% 5. evaluation
%---------------------------------
Great efforts have been made by using Deep Neural Networks to recognize multi-label images. Since multi-label image classification is very complicated, people seek to use the attention mechanism to guide the classification process.
However, conventional attention-based methods always analyzed images directly and aggressively. It is difficult for them to well understand complicated scenes.
In this paper, we propose a global/local attention method that can recognize an image from coarse to fine by mimicking how human-beings observe images. Specifically, our global/local attention method first concentrates on the whole image, and then focuses on local specific objects in the image.
We also propose a joint max-margin objective function, which enforces that the minimum score of positive labels should be larger than the maximum score of negative labels horizontally and vertically. This function can further improve our multi-label image classification method. We evaluate the effectiveness of our method on two popular multi-label image datasets (i.e., Pascal VOC and MS-COCO). Our experimental results show that our method outperforms state-of-the-art methods.
\end{abstract}


%\section{Introduction}
%This document is a model and instructions for \LaTeX.
%Please observe the conference page limits. 
\section{Introduction}

%---------------------------------
%1) what is multi-label classification
%2) traditional methods
%3) attention mechanism
%4) our works
%5) contributions
%---------------------------------
In a smart city\cite{cocchia2014smart,su2011smart}, multi-label scenes are much common, and accurately recognizing multiple label is quite important. 
For example, by recognizing every traffic routes and analyzing flows through monitors, a smart city is able to ease traffic jams.
Recently, some study about multi-label image classification in smart cities are draw attention of researchers\cite{sanghi2017automatic,alhamoud2016activity}.
Multi-label image classification seeks to recognize all possible objects/labels in a given image. %is a more difficult task than single-label one.
Because of the dramatic development of deep learning and the availability of large-scale datasets such as ImageNet~\cite{deng2009imagenet}, there exist many studies on single-label image classification~\cite{simonyan2014very,he2016deep}.
However, scenes around us are always with multiple objects/labels. 
Unfortunately, multi-label image classification are more difficult than single-label one since the complicated structure and the internal label dependencies.

Recently, methods based on Deep Neural Networks become popular.
On the one hand, due to the success of Convolutional Neural Networks (CNNs) on single-label image classification, a large number of methods directly apply CNNs to multi-label tasks\cite{gong2013deep,NIPS2012_4824,sharif2014cnn,wei2014cnn}.
%Gong \etal \cite{gong2013deep} compare several multi-label loss functions and use a CNN model that similar to AlexNet~\cite{NIPS2012_4824}. %find that a significant performance gain could be obtained by using a top-$k$ ranking objective function.
%Sharif \etal \cite{sharif2014cnn} extract features from a CNN model and leverage an extra SVM to classify the relevant labels.
%Wei \etal \cite{wei2014cnn} extract some hypotheses from source image and train them using a pre-trained CNN model.
On the other hand, some researchers additionally leverage Recurrent Neural Networks (RNNs) to model the dependencies among labels \cite{wang2016cnn,jin2016annotation,liu2016semantic}.
However, all the aforementioned works consider to indiscriminately analyze the whole image when building a multi-label image classification model, so that useless and redundant information would be equally taken into account.
For example, some blank or blur backgrounds may be behind key objects in an image are equally used in the model learning process.

\begin{figure*}[t]
	\centering\includegraphics[width=0.7\linewidth]{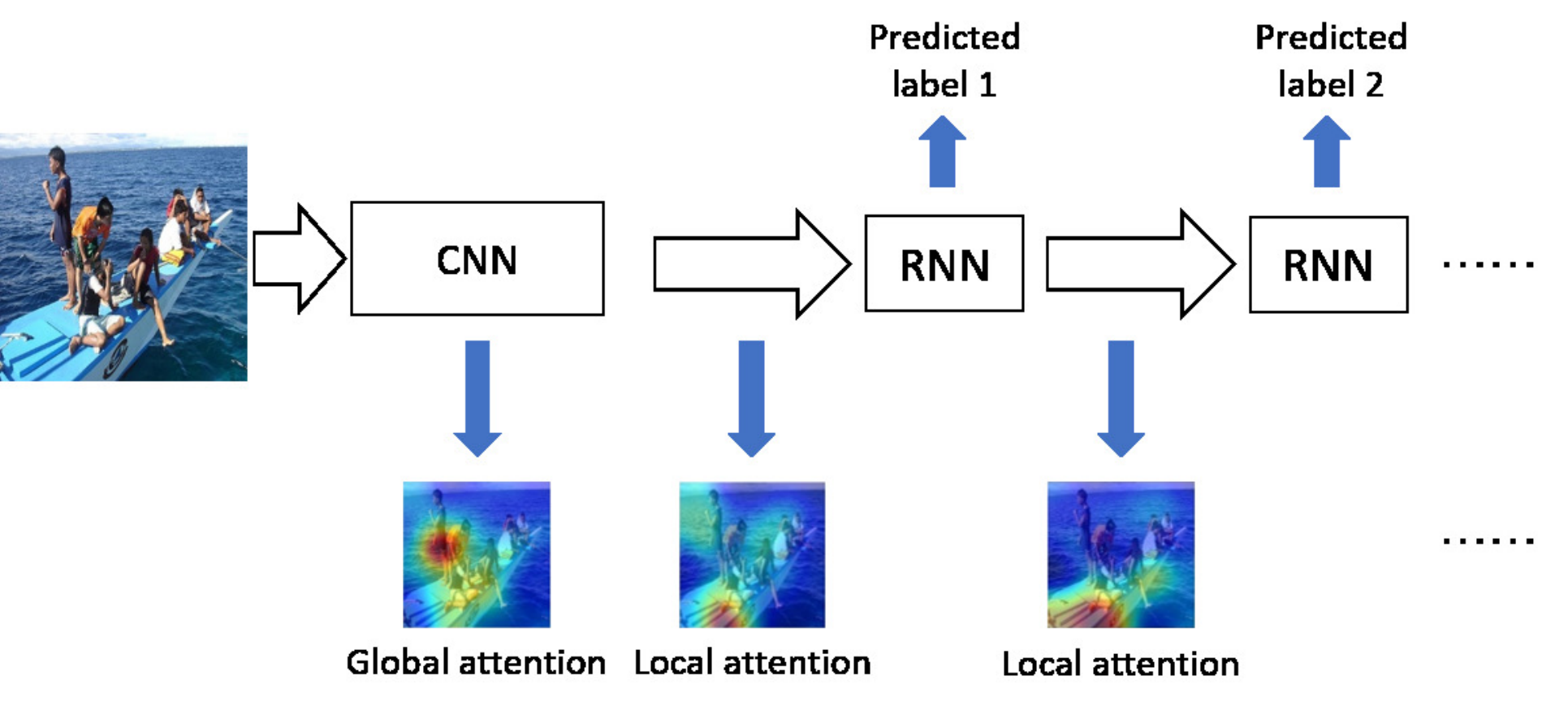}
	\caption{
		The schematic of the proposed architecture.
		The encoder CNN extracts feature from images, and computes the global attention from the feature map.
		Then, the attended features are sent to the decoder RNN.
		At every step of RNN, we focus on each local attention to guide the prediction.
%		The encoder CNN extracts ${\bm{f}}_{conv}$ and ${\bm{f}}_{fc}$ from the final convolutional layer and the final fully-connected layer, respectively.
%		For each step of LSTM, we use the hidden state and ${\bm{f}}_{conv}$ to compute the dynamic attentive context $\bm{z}_t$, which is described in Section \ref{sec:visatt} detailedly.
%		We also merge the image embedding computed from ${\bm{f}}_{fc}$ with the output of LSTM every step.
	}\label{fig:architecture}
\end{figure*}

In this paper, inspired by the success of the attention mechanism~\cite{bahdanau2014neural,xu2015show,zhou2017global,you2016image},%li2015seg,yuan2017sift,zhou2016copy,xu2016ask,lu2016hierarchical,wang2016vqa},
%Recently, the attention mechanism becomes popular, because it helps machines to focus on certain parts of images. 
%Since observers could have different knowledge background, different strategies, or different goals when they look at images, they could see different things inside images.
%The attention mechanism was first described in human attention mechanism theories~\cite{rensink2000dynamic}, and was recently adapted for Neural Machine Translation~\cite{bahdanau2014neural}, Image Captioning~\cite{xu2015show,zhou2017global,you2016image}, Image Retrieval~\cite{xia2016ir,zhou2017dup,wang2017feature}, Information Forensics and Security~\cite{li2015seg,yuan2017sift,zhou2016copy} and Visual Question Answering~\cite{xu2016ask,lu2016hierarchical,wang2016vqa}.
%%The vanilla neural networks always observe a scene in general and gather high-level information.
%However, few works used the attention mechanism to address the multi-label image classification problem.
%\cite{zhu2017learning} proposed to learn semantic and spatial relations jointly and generate attention maps for all labels.
%%Wang \etal \cite{bibid} propose to transform the attention problem into an attentional regions discovering one.
%%However, their work compute attention for all labels, which may result in a large number of additional parameters.
%However, their work computed attentions for all labels spatially, which is too direct and aggressive.
we propose a global/local attention method to for multi-label image classification that can classify images from coarse to fine.
The model can imitate how human beings observe a scenery image---they first observes the image with a global attention to find the areas that may have objects, and then focuses on these areas to consider what object is inside each area. 
The process is simply shown in Fig.~\ref{fig:architecture}. 
The global attention, which is generated from the final convolutional layer in CNN, denotes a general attentive area, \ie, an overview of an image.
Then, we generate local attention in every step of RNN, which denotes each specific attentive area for each predicted label.
%Thus, we can learn two local areas (i.e., ``person" and ``aeroplane") from the image shown in Figure~\ref{fig:example} by the global attention, and then we can further predicts the corresponding labels with these two local attentions.
Additionally, we propose a joint max-margin objective function to separate the positive and negative prediction in the time domain, which can effectively improve the performance.
We evaluate our method on two popular multi-label image datasets, and the experimental results show that our method is better than the other state-of-the arts.

%The contributions of this paper are as follows:
%
%\begin{itemize}
%	\item A global/local attention mechanism, observing the scene from coarse to fine, is introduced to help classify multi-label images sequentially.
%	\item A joint max-margin objective function is proposed to separate the positive and negative prediction in the time domain.
%\end{itemize}
%
%The rest of the paper is structured as follows. Section~\ref{sec:ml} discusses popular existing multi-label image classification methods. Section~\ref{sec:method} describes the details of our proposed method. Section~\ref{sec:exp} investigates the performance of our proposed method through a series of experiments. Finally, the conclusions of this paper are given in Section~\ref{sec:conclusion}. 
\section{Related Work}\label{sec:ml}

%-------------------------------------------
% 1. multi-label problem
% 2. attention mechansim
% 2. multi-label with attention
%-------------------------------------------

\subsection{Multi-label image classification}

%--------------------------------------------
% 1. what is  ml image classification
% 2. the traditional methods
% 3. the deep method
% 4. the drawback
%--------------------------------------------

Multi-label classification is with wide applications in many areas, especially for image classification, and lots of efforts have been made for this task.
Traditional methods can be decomposed into two categories \cite{zhang2014review}, \ie the problem transformation \cite{boutell2004learning,read2009classifier,read2011classifier} and algorithm adaptation \cite{zhang2007ml,clare2001knowledge}.
Recently, methods based on CNNs become popular in single-label image classification for its strong capability in learning discriminative features.
Some researchers attempted to directly apply CNNs on multi-label image classification.
Gong \etal\cite{gong2013deep} built a CNN architecture similar to \cite{NIPS2012_4824} to tackle this problem, and trained a CNN model with top-\textit{k} ranking objectives.
Wei \etal\cite{wei2014cnn} fine-tuned the network pre-trained on ImageNet~\cite{deng2009imagenet} with the squared loss for multi-label image classification (I-FT).
Some works employed an object detection framework to strengthen the performance of CNNs.
For example, Wei \etal provided a regional solution that allows predicting labels independently at the regional level (H-FT).
Some approaches use RNNs to model label dependencies.
Wang \etal\cite{wang2016cnn} utilized CNNs to extract image features, and then utilized RNNs to model correlations among labels.
In \cite{wang2016cnn}, the authors combined the image embedding with the output of Long Short-Term Memory (LSTM) every step, and then passed the combined vector to the final fully connected layer to predict the current label.
Liu \etal\cite{liu2016semantic} regularized CNN by ground truth semantic concepts, and then used the prediction to set the LSTM initial states. 
%These CNN-RNN based methods benefit by the development of image captioning \cite{vinyals2015show,xu2015show,mao2014explain,wu2016image}.
Although the performance of multi-label classification has been significantly improved by using CNNs and RNNs, these methods always consider to extract features from the whole image. This results in that much redundant information would be equally considered in the multi-label classification model training process. In fact, relevant objects may be only little parts of an image. Some researchers started to leverage the attention mechanism to guide multi-label classification.

\subsection{Attention mechanism}

The attention mechanism forces a learning model to focus on relevant parts of an original data.
%It is firstly applied to the machine translation, then followed by some image captioning models.
Bahdanau \etal\cite{bahdanau2014neural} proposed a model to search a set of possible positions while generating the target word in Neural Machine Translation.
This mechanism was then applied to the research field that combines vision and language.
In \cite{xu2015show}, Xu \etal took hard and soft attention-based methods to generate image descriptions.
You \etal\cite{you2016image} ran a set of attribute detectors to get a list of visual attributes and fuse them into the RNN hidden state.
Lu \etal\cite{lu2016hierarchical} proposed a co-attention model that combines the language information and the image information in the task of Visual Question Answering.
With the attention mechanism, the model can learn the attention by itself, which can intuitively guide the model to observe the data.
%The attention is learned by the model itself, which can intuitively guide the model to observe the data.
However, few works applied this mechanism to the multi-label image classification.
Zhu \etal\cite{zhu2017learning} proposed to learn semantic and spatial relations jointly and generate attention maps for all labels.
Although their work computed attentions for all labels, this may also result in a large number of additional parameters.

In this paper, we argue that the attention can also be learned from coarse to fine.
Almost all existing attention-based methods analyzed the whole image directly and cursorily, and we think this should follow a progress. 
When coming across a complicated scene, %it is difficult for us to focus on specific objects at the beginning.
we need to look around in general firstly and then search specific objects one by one. 
Therefore, we propose a global-local attention method for multi-label image classification. The details of our proposed method will be explained in Section \ref{sec:method}.
%However, \cite{bibid} proposes that the attention mechanism is weak for the short sequence.
%As we know, the result of multi-label classification are short sequences.

%In CNN-RNN based methods, the labels can be treated as a denser sequence than a sentence because they has no syntactic structure and no unnecessary qualifiers.
%In this paper, we on top of this premise and design a dynamic attention mechanism on multi-label classification.
%This is different from the long sentence sequence, and we will explain that in Section \ref{sec:method}.

\section{Method}\label{sec:method}

%In this section, we illustrate our model.

\subsection{Problem definition}

Multi-label classification is to predict all possible labels for an image.
Given a set of images $\mathcal{X}=\{\bm{x}_1,\bm{x}_2,\dots,\bm{x}_N\}$ and their corresponding labels $\mathcal{Y}=\{\bm{y}_1,\bm{y}_2,\dots,\bm{y}_N\}$, where $N$ is the number of images in the set, our work is to learn a hypothesis $h:\mathcal{X}\rightarrow\mathcal{Y}$ that maps an input image $\bm{x}$ to output $\bm{y}$.
%We take an image as the input and use our model to predict all the possible labels involved in the image.
For the \textit{i}-th image $\bm{x}_i$, we denote the corresponding labels as $\bm{y}_i=\{y_{i1},y_{i2},\dots,y_{iC}\}$, where $y_{ij}=1$ means the image $\bm{x}_i$ is labeled with label $j$ while $y_{ij}=0$ is on the contrary. $C$ is the number of possible labels.
%Here, $\bm{Y}_i$ has already been ordered by the frequency of all labels in dataset.
%Here, the number $D$ is  fixed and known in advance, whilst the number $C$ is unknown and different when predicting different images.
%Our model can also decide the value of $C$ by itself, while the traditional methods \cite{gong2013deep,sharif2014cnn,oquab2014learning,wei2014cnn} often predict multiple labels with a threshold or pick the top-\textit{k} ranking labels.
%In the traditional methods (such as \cite{gong2013deep,sharif2014cnn,oquab2014learning,wei2014cnn}), they often determine the number of labels (i.e. the number $C$) based on some threshold or by picking the top-\textit{k} ranking labels, where $k$ is a user predefined number.
%On the contrary, we will propose a new scheme which seeks to learn  the parameter $C$ automatically.

\subsection{The framework of our model}

Our overall model follows the \textit{encoder-decoder} design pattern \cite{pan2016encoder}, which transforms data from one representation to another.
In the proposed model, the encoder is a VGG-16 model \cite{simonyan2014very}, which has been proved to extract features from image effectively.
From the VGG-16 model, we extract two types of features from each image.
The first type of features comes from the final convolutional layer, presenting the structural information of an image and denoting as $\bm{f}_{conv}=\{\bm{a}_1,\bm{a}_2,\dots,\bm{a}_L\}$, where $L$ is the number of regions in the feature map.
The other type of features is from the last fully-connected layer, including more higher-level information of an image and denoting as $\bm{f}_{fc}$.
The decoder is an RNN model. 
In this paper, we used Long Short-Term Memory \cite{hochreiter1997long} (LSTM).
LSTM adds three extra gates to the vanilla RNN, \ie, the input gate, the forget gate and the output gate.
%As shown in Fig.~\ref{fig:lstm},
%the input gate controls the input data it should read;
%the forget gate controls the previous state it should forget;
%and the output gate controls the current state it should output.
Following \cite{wang2016cnn}, multiple labels in multi-label classification can be regarded as a sequence, and the RNN decoder is used to recognize each specific object one by one.
%Following \cite{zaremba2014learning}, the forward passing at step $t$ can be defined as
%
%\begin{align}
%	\bm{g}_t&=\tanh(\bm{W}_{xc}\bm{y}_t+\bm{W}_{hc}\bm{h}_{t-1}+\bm{b}_c),\\
%	\bm{i}_t&=\sigma(\bm{W}_{xi}\bm{y}_t+\bm{W}_{hi}\bm{h}_{t-1}+\bm{b}_i),\\
%	\bm{f}_t&=\sigma(\bm{W}_{xf}\bm{y}_t+\bm{W}_{hf}\bm{h}_{t-1}+\bm{b}_f),\\
%	\bm{o}_t&=\sigma(\bm{W}_{xo}\bm{y}_t+\bm{W}_{ho}\bm{h}_{t-1}+\bm{b}_o),\\
%	\bm{c}_t&=\bm{f}_t\odot \bm{c}_{t-1}+\bm{i}_t\odot \bm{g}_t,\\
%	\bm{h}_t&=\bm{o}_t\odot\tanh(\bm{c}_t),\\
%	\bm{p}_{t}&=\text{Softmax}(\bm{h}_t),
%\end{align}

%where all $\bm{W}$-s and $\bm{b}$-s are trainable weights and bias.
%$\bm{y}_t$ denotes the input label in step $t$, and $\bm{h}_{t-1}$ is the last hidden state.

\subsection{Visual attention mechanism}

In this section, we describe our visual attention mechanism.
In our model, we leverage two types of attentions, \ie, the global attention and the local attention.
For the global attention, a more general attentive area is highlighted, while a more fine-grained one is highlighted for local attention.

\subsubsection{Global attention}

Our global attention $\hat{\bm{\alpha}}=\{\hat{\alpha}_1,\hat{\alpha}_2,\dots,\hat{\alpha}_N\}$ is computed from $\bm{f}_{conv}$.
For the $i$-th region, it corresponds to a positive weight

%\begin{figure}
%	\centering\includegraphics[width=0.6\columnwidth]{figs/global.png}
%	\caption{The architecture to obtain global attention.}\label{fig:global}
%\end{figure}

\begin{figure*}
	\centering
	\subfigure[Global attention]{\centering\includegraphics[width=.4\linewidth]{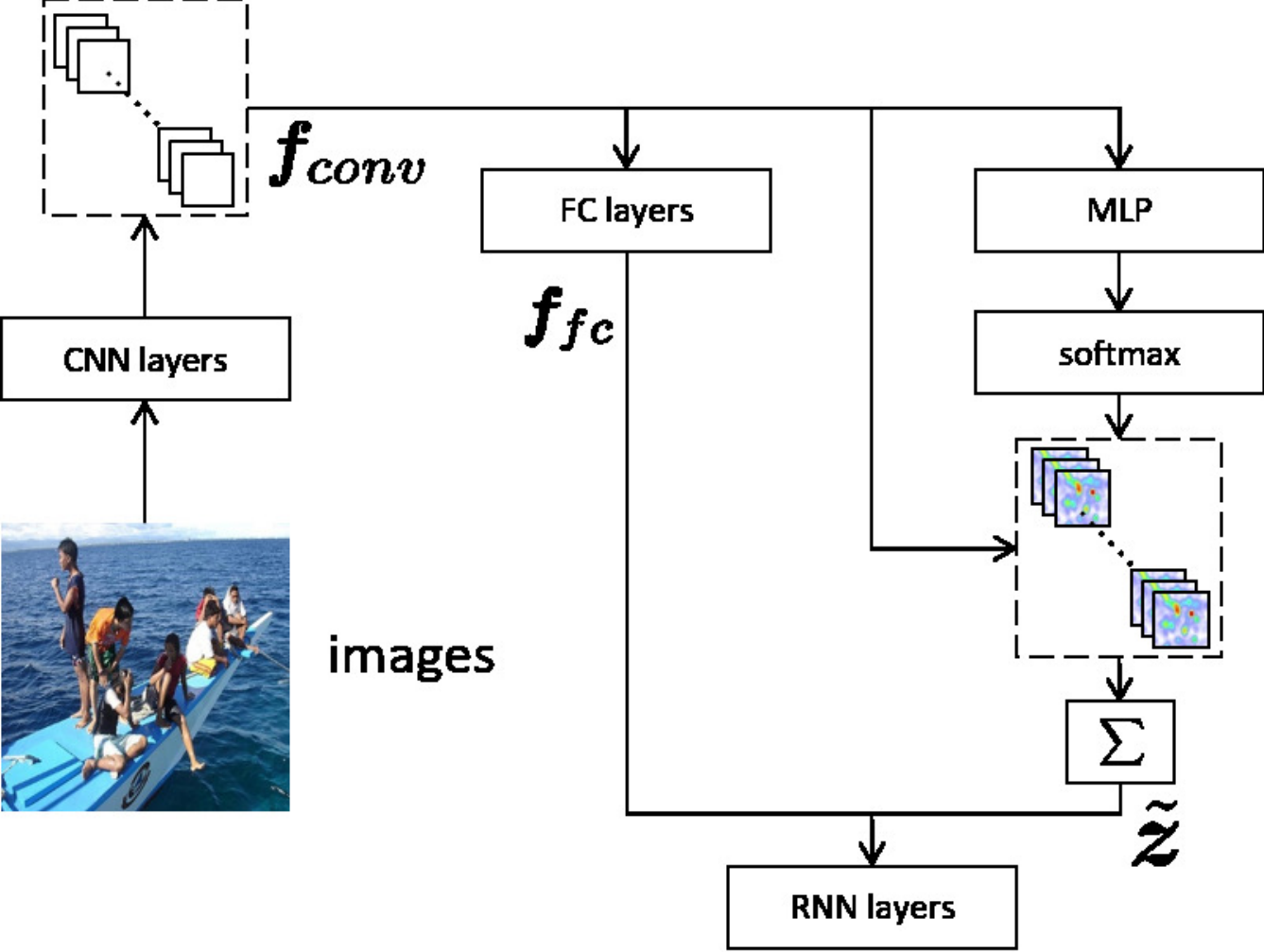}\label{fig:global}}
	\subfigure[Local attention]{\centering\includegraphics[width=0.4\linewidth]{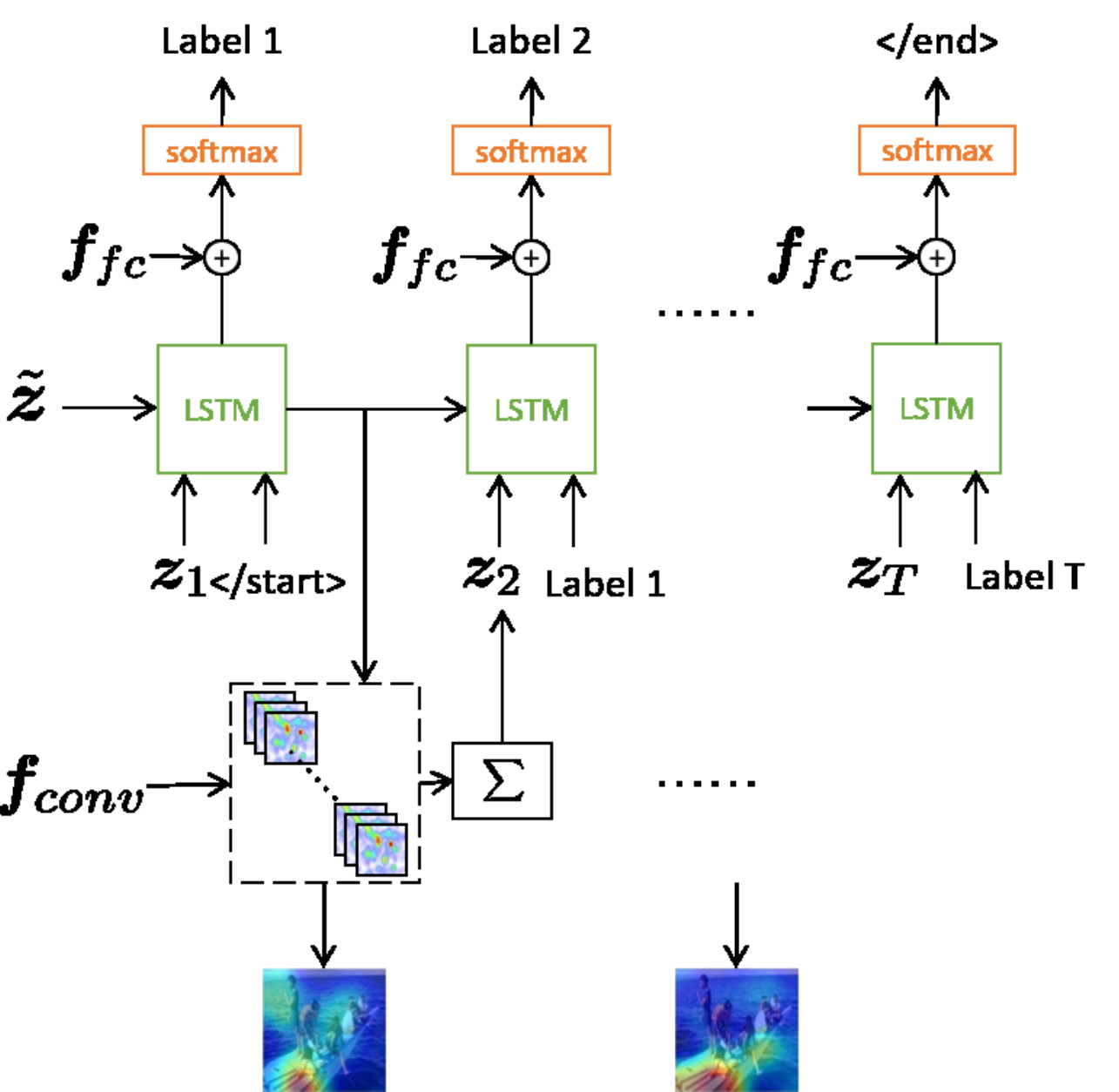}\label{fig:local}}
	\caption{The architecture to obtain global and local attention.}%\label{fig:global}
\end{figure*}

\begin{equation}
%	\hat{\alpha}_{i}=\frac{\exp(\hat{e}_{i})}{\sum_{k=1}^{N}\exp(\hat{e}_{k})},~i=\{1,2,\dots,N\}
	\hat{\alpha}_i = \text{softmax}(\tanh(\bm{W}_g\bm{a}_i+\bm{b}_g)),
\end{equation}
where $\hat{\alpha}_i$ is a scalar presenting the degree of the $i$-th region's importance. %, and $\hat{\bm{e}}=\{\hat{e}_{1},\hat{e}_{2},\dots,\hat{e}_{N}\}$ represents the energy of the weight $\hat{\bm{\alpha}}$.
%That is, for each region of feature map $\bm{f}_{conv}$, $\hat{\alpha}$ owns a corresponding weight to represent the degree of its importance.
With $\hat{\bm{\alpha}}$, we can compute the excepted aligned global context $\hat{\bm{z}}$, and the process can be shown in Fig.~\ref{fig:global}. %by the sum of all weighted $\bm{a}_i$.

\begin{equation}
\hat{\bm{z}}=\sum_{i=1}^N\hat{\alpha}_{i}\bm{a}_i.
\end{equation}

Note that we used the sum of all weighted $\bm{a}_i$ to compute the expected aligned contexts.
%In the machine translation, this model scores the contribution of each input when predicting an output word.
%The sequence of a sentence lies in the 1-D space. However, each image is at least 2-D.
This attentive context shows how the weight influences the feature maps.%, so we regard their set $\bm{z}_t$ as another special features and feed them into LSTM as the next input.
Thus, unlike traditional sequence learning that zero-initializing LSTM, in our architecture, we initialized it with the average of $\hat{\bm{z}}$.

\begin{equation}
\bm{c}_0=I_c(\hat{\bm{z}}/L),~\bm{h}_0=I_h(\hat{\bm{z}}/L),
\end{equation}

The initialization of parameters is quite important.
\cite{yang2016review} considered that the attention mechanism lacks global modeling abilities in common sequential learning.
Initializing the memory cell and the hidden state in this way helps LSTM learn the whole non-attentive feature maps and a glance to the original image.
Moreover, because $\hat{\bm{z}}$ is an attentive global context, our model can first give a general area that may contain some meaningful objects.

%\begin{equation}
%	\hat{e}_{i} = \tanh(\bm{W}_g\bm{a}_i+\bm{b}_g)	
%\end{equation}

\subsubsection{Local attention}

After the proposed model has observed an image in a general way, we expect our model like human beings to focus on every specific object. Therefore, as shown in Fig.~\ref{fig:local}, our model computes local attention at each step of LSTM.

For all regions at step $t$, similar to the global attention, we used a positive weight $\bm{\beta}_{t}=\{\beta_{t1},\beta_{t2},\dots,\beta_{tN}\}$ to decide which location is the right attentive place for the next label. Its element $\beta_{ti}$ is computed by

\begin{equation}
%\beta_{ti}=\frac{\exp(e_{ti})}{\sum_{k=1}^{N}\exp(e_{tk})}.
\beta_{ti}=\text{softmax}({g(\bm{a}_i,\bm{h}_{t-1})}).
\label{beta_ti}
\end{equation}

%\begin{figure}
%	\centering\includegraphics[width=0.6\columnwidth]{figs/lstm.pdf}
%	\caption{An architecture of LSTM with a dynamic attentive context $\bm{z}_t$.}\label{fig:lstm}
%\end{figure}

In Eq.~(\ref{beta_ti}), $\beta_{ti}$ presents the prior hidden state $\bm{h}_{t-1}$,
%\begin{equation}
%e_{ti}=g(\bm{a}_i,\bm{h}_{t-1}).
%\label{e_t}
%\end{equation}
$g$ is a simple Multi-Layer Perceptron, which reflects the importance of the feature $\bm{a}_i$ as well as the hidden state $\bm{h}_{t-1}$ and decides the next state of LSTM.
Therefore, LSTM is forced to pay more attention to these regions with larger weights. % of the original image.
Then, we can compute the dynamic context $\bm{z}_t$ as follows.

\begin{equation}
\begin{aligned}
\bm{z}_t=\beta_t\cdot\phi(\{\bm{a}_i\},\bm{\alpha}_{t})=\beta_t\sum_{i=1}^N\alpha_{ti}\bm{a}_i.
\end{aligned}
\label{z_t}
\end{equation}

We treated their set $\bm{z}_t$ as another special features and feeded them into LSTM as the next input.
%In a conventional model, the input of LSTM is usually the label embedding ${\bm x}_t$, computed from the ordered label sequence, and the previous hidden state $\bm{h}_{t-1}$.
%For convenience, we concatenate the label embedding and the dynamic context $\bm{z}_t$ as the uniform input of LSTM.
That means for every step of the LSTM's recurrence, the model must take the possible area into account and overlook some unimportant information.
As a consequence, following \cite{zaremba2014learning}, the forward passing at step $t$ can be defined as follows.

\begin{align}
\bm{g}_t&=\tanh(\bm{W}_{xc}\bm{y}_t+\bm{W}_{hc}\bm{h}_{t-1}+\bm{W}_{zc}\bm{z}_t+\bm{b}_c),\\
\bm{i}_t&=\sigma(\bm{W}_{xi}\bm{y}_t+\bm{W}_{hi}\bm{h}_{t-1}+\bm{W}_{zi}\bm{z}_t+\bm{b}_i),\\
\bm{f}_t&=\sigma(\bm{W}_{xf}\bm{y}_t+\bm{W}_{hf}\bm{h}_{t-1}+\bm{W}_{zf}\bm{z}_t+\bm{b}_f),\\
\bm{o}_t&=\sigma(\bm{W}_{xo}\bm{y}_t+\bm{W}_{ho}\bm{h}_{t-1}+\bm{W}_{zo}\bm{z}_t+\bm{b}_o),\\
\bm{c}_t&=\bm{f}_t\odot \bm{c}_{t-1}+\bm{i}_t\odot \bm{g}_t,\\
\bm{h}_t&=\bm{o}_t\odot\tanh(\bm{c}_t),
\end{align}
where all $\bm{W}$-s and $\bm{b}$-s are trainable weights and biases.
$\bm{y}_t$ denotes the input label in step $t$, and $\bm{h}_{t-1}$ is the last hidden state.
%The LSTM model in our architecture is shown in Fig ~\ref{fig:lstm}.

%\begin{figure}[t]
%	\centering\includegraphics[width=0.6\columnwidth]{figs/local.pdf}
%	\caption{An architecture of LSTM to obtain local attention.}
%	\label{fig:local}
%\end{figure}

%In Equation \ref{z_t}, we also add a selective gate $\beta_t$ to the dynamic context $\bm{z}_t$.
%The gate controls of whether the current attentive context should consider more about the correlation with the previous labels.
%It is computed by the current hidden state:
%
%\begin{equation}
%\beta_t=\sigma(\bm{W}_{\beta t}\bm{h}_{t-1}+\bm{b}_{\beta t}).
%\end{equation}

\subsection{Objective function}

\subsubsection{Horizontal max-margin objective}

We obtained a prediction at every step for the $i$-th image. Therefore, we will obtain a set of predictions $\bm{P}_i=\{\bm{p}_{i1},\bm{p}_{i2},\dots,\bm{p}_{iT}\}$ at the end of the sequence.
For the prediction at the $t$-th step $\bm{p}_{it}$, it is a vector with length $L$, where $L$ is the number of all classes.
Actually, we obtained the final prediction $\hat{\bm{p}}_i=\{p_{i}^j,p_{i}^j,\dots,p_{i}^j\}$ by a max-pooling for each class. For the $j$-th class, we have

\begin{equation}
p_{i}^j=\text{max}(p_{i1}^j,p_{i2}^j,\dots,p_{iT}^j).
\end{equation}

To separate the positive and negative prediction, we assumed that a max margin is between the minimum positive and maximum negative prediction. That is,

\begin{equation}
%	f_{\min}^+(\bm{x}_i)>f_{max}^-(\bm{x}_i)+\epsilon
\text{min}_+(\hat{\bm{p}}_{i})>\text{max}_-(\hat{\bm{p}}_{i})+\epsilon
\end{equation}

where $\text{min}_+(\hat{\bm{p}}_{i})$ and $\text{max}_-(\hat{\bm{p}}_{i})$ mean the minimum positive and the maximum negative prediction respectively.
$\epsilon$ is the joint max-margin, and is pre-defined before training.
As a result, we have a constrain for the prediction as follows.

\begin{equation}
R_1 = \sum_{i=1}^N\text{max}(\text{min}_+(\hat{\bm{p}}_{i})-\text{max}_-(\hat{\bm{p}}_{i})+\epsilon,0)
\end{equation}

\begin{table*}[t]\scriptsize
	\centering  %???
	\caption{Experimental Results on Pascal VOC 2007.}
	%	\begin{tabular}{l|p{.34cm}p{.34cm}p{.34cm}p{.34cm}p{.34cm}p{.34cm}p{.34cm}p{.34cm}p{.34cm}p{.34cm}p{.34cm}p{.34cm}p{.34cm}p{.34cm}p{.34cm}p{.34cm}p{.34cm}p{.34cm}p{.34cm}p{.34cm}|l}	
	
	\resizebox{\linewidth}{!}{
		\begin{tabular}{l|cccccccccccccccccccc|l}
			%	\begin{tabular}{l|cccccccccccccccccccc|l}
			\hline			
			& plane & bike & bird & boat & bottle & bus & car & cat & chair & cow & table & dog & horse & motor & person & plant & sheep & sofa & train & tv & mAP\\
			\hline\hline
			INRIA & 77.2 & 69.3 & 56.2 & 66.6 & 45.5 & 68.1 & 83.4 & 53.6 & 58.3 & 51.1 & 62.2 & 45.2 & 78.4 & 69.7 & 86.1 & 52.4 & 54.4 & 54.3 & 75.8 & 62.1 & 63.5\\
			FV    & 75.7 & 64.8 & 52.8 & 70.6 & 30.0 & 64.1 & 77.5 & 55.5 & 55.6 & 41.8 & 56.3 & 41.7 & 76.3 & 64.4 & 82.7 & 28.3 & 39.7 & 56.6 & 79.7 & 51.5 & 58.3\\
			CNN-SVM  		& 88.5 & 81.0 & 83.5 & 82.0 & 42.0 & 72.5 & 85.3 & 81.6 & 59.9 & 58.5 & 66.5 & 77.8 & 81.8 & 78.8 & 90.2 & 54.8 & 71.1 & 62.6 & 87.4 & 71.8 & 73.9\\
			I-FT			& 91.4 & 84.7 & 87.5 & 81.8 & 40.2 & 73.0 & 86.4 & 84.8 & 51.8 & 63.9 & 67.9 & 82.7 & 84.0 & 76.9 & 90.4 & 51.5 & 79.9 & 54.3 & 89.5 & 65.8 & 74.5\\
			HCP-1000C  		& 95.1 & 90.1 & 92.8 & 89.9 & 51.5 & 80.0 & 91.7 & 91.6 & 57.7 & 77.8 & 70.9 & 89.3 & 89.3 & 85.2 & 93.0 & 64.0 & 85.7 & 62.7 & 94.4 & 78.3 & 81.5\\
			HCP-2000C  		& 96.0 & 92.1 & 93.7 & 93.4 & 58.7 & 84.0 & \textbf{93.4} & 92.0 & 62.8 & \textbf{89.1} & 76.3 & 91.4 & 95.0 & 87.8 & 93.1 & \textbf{69.9} & 90.3 & 68.0 & 96.8 & 80.6 & 85.2\\
			CNN-RNN  		& 96.7 & 83.1 & 94.2 & 92.8 & \textbf{61.2} & 82.1 & 89.1 & \textbf{94.2} & 64.2 & 83.6 & 70.0 & \textbf{92.4} & 91.7 & 84.2 & 93.7 & 59.8 & \textbf{93.2} & \textbf{75.3} & \textbf{99.7} & 78.6 & 84.0\\
			\hline
			VGG				& 96.5 & 89.5 & 92.8 & 92.8 & 58.0 & \textbf{85.6} & 90.0 & 92.5 & 70.3 & 85.2 & 76.0 & 91.7 & 89.9 & 87.0 & 94.3 & 68.8 & 87.1 & 68.3 & 97.3 & 80.7 & 84.7\\
			%CNN-RNN Inject  & 96.1 & 90.6 & 92.8 & 92.2 & 51.6 & 79.5 & 88.8 & 90.8 & 70.5 & 82.1 & 72.4 & 90.8 & 94.0 & 87.8 & 95.0 & 64.7 & 87.6 & 64.5 & 97.1 & 77.0 & 83.3\\
			VGG+LSTM   		& 96.7 & 91.1 & 93.8 & 92.7 & 55.4 & 81.7 & 90.2 & 91.0 & 67.6 & 83.9 & 75.4 & 91.7 & 94.3 & 90.8 & 94.8 & 65.5 & 86.6 & 67.5 & 97.1 & 79.4 & 84.4\\
			VGG+LSTM+L 		& 97.0 & \textbf{92.5} & 93.8 & 93.3 & 59.3 & 82.6 & 90.6 & 92.0 & \textbf{73.4} & 82.4 & 76.6 & \textbf{92.4} & 94.2 & \textbf{91.4} & 95.3 & 67.9 & 88.6 & 70.1 & 96.8 & 81.5 & 85.2 \\
			VGG+LSTM+L/G 	& \textbf{97.5} & 91.0 & 94.2 & \textbf{95.0} & 56.2 & 84.9 & 90.9 & 92.3 & 71.6 & 86.8 & 76.2 & 92.0 & 94.6 & 89.5 & 95.5 & 66.4 & 86.2 & 70.0 & 96.9 & 81.4 & 85.4 \\
			VGG+LSTM+L/G+MM	& 97.0 & 92.4 & \textbf{94.4} & 93.6 & 59.1 & 83.8 & 90.7 & 92.5 & 69.8 & 84.2 & \textbf{76.9} & 91.8 & \textbf{95.1} & 91.1 & \textbf{96.2} & 66.2 & 88.0 & 69.2 & 96.2 & \textbf{82.6} & \textbf{85.6} \\
			%Ours-dependency	& \textbf{97.2} & 92.0 & 94.0 & 92.9 & 57.8 & 84.2 & 91.1 & \textbf{93.3} & 70.8 & 82.6 & \textbf{76.6} & \textbf{92.9} & 94.4 & 90.5 & \textbf{95.7} & 67.0 & 86.3 & 69.7 & 97.6 & \textbf{83.0} & 85.5 \\
			\hline
	\end{tabular}}%??
	\label{tab:voc}
\end{table*}

\begin{figure*}[t]
	\centering\includegraphics[width=.8\linewidth]{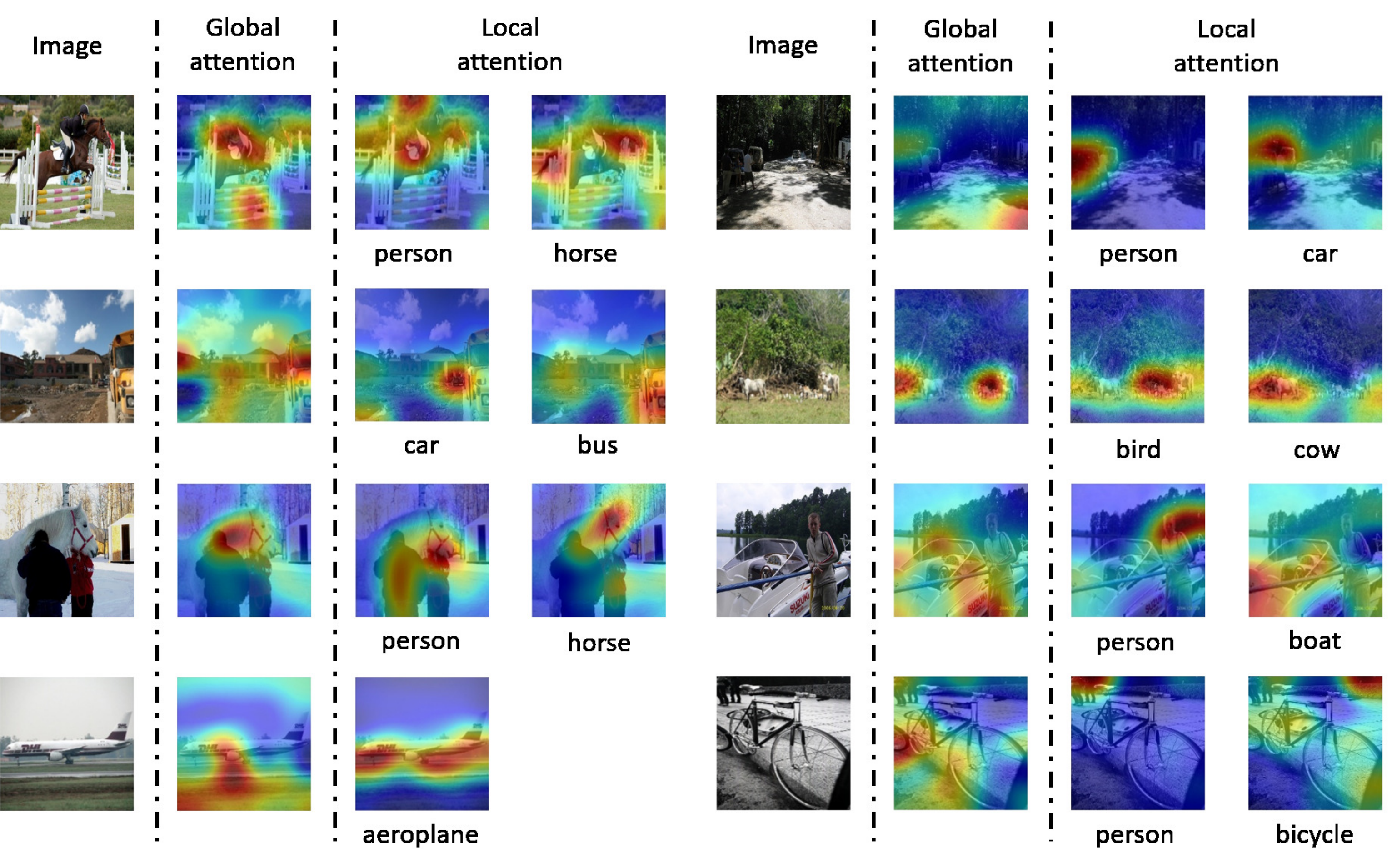}
	\caption{Visualization of the attentive images on PASCAL VOC 2007.}
	\label{fig:att_vis-1}
\end{figure*}

\begin{figure*}[t]
	\centering\includegraphics[width=.8\linewidth]{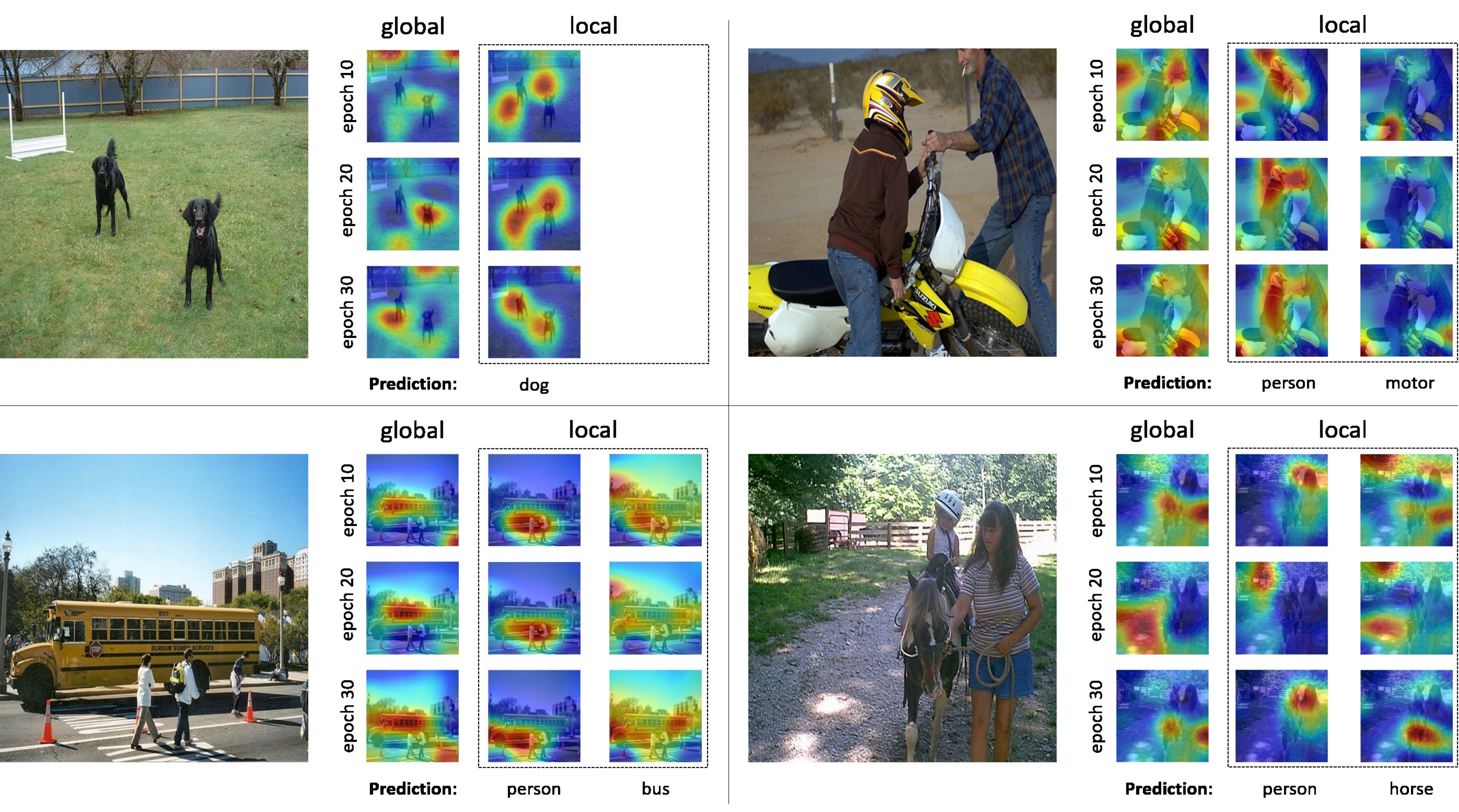}
	\caption{Visualization of attentive images on PASCAL VOC 2007 for every 10 epochs.}
	\label{fig:att_vis-2}
\end{figure*}

\begin{table*}[t]\small
	\centering
	\caption{Experimental Results on MS-COCO.}
	\resizebox{\linewidth}{!}{
		\begin{tabular}{l|llllllll|l}
			\hline			
			Method & P@3/5(\%)$\uparrow$ & R@3/5(\%)$\uparrow$ & F@3/5(\%)$\uparrow$ & H@3/5$\downarrow$ & A@3/5(\%)$\uparrow$ & 1-err(\%)$\downarrow$ & C$\downarrow$ & rloss$\downarrow$ & mAP(\%)$\uparrow$\\
			\hline\hline
			%		CNN Multi-label & \textbf{59.57}/42.74 & \textbf{73.04}/\textbf{82.14} & \textbf{65.02}/\textbf{56.22} & 0.029/0.045 & \textbf{47.14}/37.56 & \textbf{10.41} & 49.38 & 0.028 & 64.41\\	
			CNN \cite{wang2016cnn}   		& -/- & -/- & -/- & -/- & -/- & - & - & - & 57.20\\
			CNN-RNN \cite{wang2016cnn}   		& -/- & -/- & -/- & -/- & -/- & - & - & - & 61.20\\
			VGG & 58.78/42.23 & \textbf{71.96}/81.14 & 64.70/55.55 & 0.029/0.046 & 46.25/36.98 & \textbf{11.25} & \textbf{49.31} & 0.032 & 63.94\\
			VGG+LSTM   & 57.92/41.18 & 70.87/79.24 & 63.74/54.19 & 0.030/0.047 & 45.32/35.79 & 12.12 & 47.81 & 0.050 & 61.54\\	
			VGG+LSTM+L		& 58.76/42.55 & 71.67/81.45 & 64.58/55.90 & 0.029/0.045 & 46.19/37.33 & 12.33 & 49.80 & 0.031 & 63.93\\
			VGG+LSTM+L/G	& 58.98/42.62 & 71.86/81.43 & 64.79/55.95 & 0.029/0.045 & 46.40/\textbf{37.52} & 11.88 & 49.83 & \textbf{0.029} & 64.07\\
			VGG+LSTM+L/G+MM	& \textbf{59.08}/\textbf{42.72} & 71.95/\textbf{81.67} & \textbf{64.88}/\textbf{56.10} & 0.029/0.045 & \textbf{46.46}/37.39 & 11.61 & 49.63 & 0.031 & \textbf{64.64}\\
			\hline
	\end{tabular}}
	\label{tab:coco}
\end{table*}

\subsubsection{Vertical max-margin objective}

With only the horizontal max-margin objective, the distance between positive and negative labels will be larger.
However, for each step, we only expect to predict one label.
Thus, even if the label is not predicted, the margin still exists.
Therefore, we proposed another vertical max-margin objective.
The prediction list  $\bm{P}_i=\{\bm{p}_{i1},\bm{p}_{i2},\dots,\bm{p}_{iT}\}$ can be regarded as a matrix $\bm{Q}\in\mathbb{R}^{T\times L}$, and
$\bm{Q}_i=[\bm{p}_{i1},\bm{p}_{i2},\dots,\bm{p}_{iT}]^\top$,
where the $t$-th row presents the prediction in step $t$ and the $l$-th column presents the $l$-th class.
Thus, for each class the minimum positive and maximum negative also have a max margin.

\begin{equation}
\text{min}_+(\bm{Q}_{i, j})>\text{max}_-(\bm{Q}_{i, j})+\epsilon_v,~~j\in{1,2,\dots,L}.
\end{equation}

where $\text{min}_+(\bm{Q}_{i, j})$ means the minimum positive prediction on class $j$ for each step, and $\text{max}_-(\bm{Q}_{i, j})$ means the maximum negative prediction.
The constrain in the vertical direction can be denoted as

\begin{equation}
R_2 = \sum_{i=1}^N\text{max}(\text{min}_+(\bm{Q}_{i, j})-\text{max}_-(\bm{Q}_{i, j})+\epsilon_v,0)
\end{equation}

%-------------------------------------------------------------------------
%Beyond that, to avoid the attention areas over-concentrated on one little point, we additionally add an attention regular term to the loss function.
%Thus, for one training instance $(\bm{x}_i,\bm{\tilde{Y}}_i)$, the loss function can be written as
%
%\begin{align}
%R_2&=\sum_{i}^{N}(1-\sum_{t}^{T}\bm{\alpha}_t)^2,\\
%R_3&=\sum_{i}^{N}(1-\hat{\bm{\alpha}})^2.
%\end{align}
%
%$R_2$ and $R_3$ represent the local and global regular term respectively.

\subsubsection{Final objective}

Although we do prediction at every step of RNN, we defined the final prediction as the max-pooling of the prediction of each step.
Formally, given a training sample $\{\bm{x}_i,\bm{y}_i\}$, we expect the model to give the prediction $\hat{\bm{y}_i}$.

We construct the final objective function as

\begin{equation}
\begin{aligned}
&\mathcal{L}(\mathcal{X},\mathcal{Y})\\=&-\frac{1}{N}\sum_{i=1}^N\sum_{j=1}^C[y_{ij}\log(\hat{y}_{ij})+(1-y_{ij})\log(1-\hat{y}_{ij})]\\
&+\lambda_1R_1+\lambda_2R_2,
\end{aligned}
\end{equation}

where $\lambda_1$ and $\lambda_2$ are the regular parameters.%, and the concentrated level is related to its value.
%$T$ is the maximal sequence length.

\section{Experiments}\label{sec:exp}

\subsection{Datasets and Experimental Settings}

We used two popular multi-label image datasets, i.e., The PASCAL Visual Object Classes Challenge (Pascal VOC)~\cite{everingham2010pascal} and Microsoft COCO (MS-COCO)~\cite{lin2014microsoft}, to evaluate our method.
Pascal VOC 2007 has 5,011 training examples and 4,952 testing examples of 20 classes.
MS-COCO dataset \cite{lin2014microsoft} has 123,287 images (82,783 training and 40,504 validating examples) of 80 different classes.

%We compare our results with that of the state-ot-the-art methods.

%-------------------------------------------------
For the proposed method, we used VGG-16 \cite{simonyan2014very} as our back-bone model of the encoder CNN.
The ${\bm{f}}_{conv}$ are extracted from the last convolutional layer \texttt{conv5\_3} and ${\bm{f}}_{fc}$ are extracted from the last fully-connected layer \texttt{fc\_7}.
The parameters of VGG-16 are pre-trained on ImageNet.
We set $\lambda_1 = 5\times10^{-2}$ and $\lambda_2 = 5\times10^{-2}$ to determine the importance of the max-margin regular term.
%$\lambda_2 = 10^{-2}$ and $\lambda_3 = 2\times10^{-2}$, that means we set the local attention twice concentrated over the global attention.
%Because we need the local attention to care more about specific objects and the global attention consider more about the overall environment.
In our experimental results, we used ``L" and ``G" to represent the model with local attention and/or global attention respectively.
And we used ``MM" to represent the model with joint max-margin objective.
We used the several common metrics (Precision[P], Recall[R], F1-Score[F], Hamming loss[H], Accuracy[A], One error[1-err], Coverage[C], Rank loss[rloss], Mean average precision[mAP]) to evaluate our method and comparison methods, and X@$k$ means metric X on top $k$.
$\downarrow$ means the lower the metric, the better the performance is, while $\uparrow$ is on the contrary.

%\begin{itemize}
%	\item Precision (P@$k$): the precision rate on top $k$;
%	\item Recall (R@$k$): the recall rate on top $k$;
%	\item F1-Score (F@$k$): the harmonic mean between precision and recall on top $k$;
%	\item Hamming loss (H@$k$): the fraction of misclassified instance-label pairs on top $k$;
%	\item Accuracy (A@$k$): accuracy on top $k$;
%	\item One error (1-err): how many times the top-ranked label is not in the set of relevant labels of examples;
%	\item Coverage (C): how far, on average, we need to go down the list of ranked labels in order to cover all the relevant labels of examples;
%	\item Rank loss (rloss): the average fraction of label pairs that are reversely ordered for particular examples;
%	\item Mean average precision (mAP): average precision is the average fraction of labels ranked higher than a particular label, and mAP is the mean average precision across all labels.
%\end{itemize}
%-------------------------------------------------
\subsection{Performance on Pascal VOC}

We first evaluated our method on Pascal VOC 2007.
The comparison to the state-of-the-art methods is shown in Table~\ref{tab:voc}.
Comparison methods include the follows:
%INRIA~\cite{harzallah2009combining}, FV~\cite{perronnin2010improving}, CNN-SVM~\cite{sharif2014cnn}, IFT, HCP~\cite{wei2014cnn} and CNN-RNN~\cite{wang2016cnn}.

\begin{itemize}
	\item INRIA~\cite{harzallah2009combining} combines object localization and image classification efficiently and makes both improved.
	\item FV~\cite{perronnin2010improving} applies Fisher Kernel~\cite{jaakkola1999exploiting} to multi-label image classification.
	\item CNN-SVM uses CNN to extract the feature of images and classifies images with SVM \cite{gu2017solution,gu2017minimax,gu2015reg}.
	\item IFT trains an AlexNet with a softmax loss.
	\item ~\cite{wei2014cnn}, based on IFT, fine-tunes the network with multiple hypotheses, and augments 1000/2000 additional classes.
	\item CNN-RNN~\cite{wang2016cnn} uses a CNN as an encoder and a RNN as a decoder, and predicts labels sequentially.
\end{itemize}

From Table~\ref{tab:voc}, We can see that our method outperforms these state-of-the-arts.
First, our \textbf{VGG+LSTM+L} that leverages the local attention has the same performance (85.2\% mAP) with \textbf{HCP-2000C}, but the latter additionally trains the model with extra 2000 classes.
Second, With global attention, our \textbf{VGG+LSTM+L/G} is better than \textbf{VGG+LSTM+L}, and reaches to 85.4\% in terms of mAP.
At last, when we use the joint max-margin objective, our \textbf{VGG+LSTM+L/G+MM} achieves the best performance (85.6\%), which shows the constructed joint max-margin objective can effectively improve the classification.

\subsection{Performance on MS-COCO}

We then evaluated our method on the dataset MS-COCO, and the experimental results are shown in Table \ref{tab:coco}.
First, our \textbf{VGG+LSTM+L/G+MM} is better than all other methods in most metrics; In terms of mAP, it reaches 64.64\% , outperforming \textbf{VGG+LSTM+L/G} (64.07\%).
Second, from Table \ref{tab:coco}, we can see that the performance of \textbf{VGG+LSTM+L/G+MM} on both top-$3$ and top-$5$ are better than that of other methods on most metrics. 
At last, the performance of \textbf{VGG} and \textbf{VGG+LSTM+L} is close, and this is probably because MS-COCO is a large dataset and the correlations among labels in it is not obvious.
For example, the label ``person" has much higher frequency than other labels. When the current prediction is ``person", it is difficult to determine which label to predict in the next step.
Another evidence is that the performance of \textbf{VGG+LSTM} is worse than that of both \textbf{VGG} and \textbf{VGG+LSTM+L}.

\subsection{Visualization of attention}\label{sec:visatt}

We visualized the attentive areas for the images on PASCAL VOC 2007 by up-sampling the attention weights with a factor of $2^4=16$ and applying a Gaussian filter.
We showed the predictions and the relative attentive areas of images in Fig.~\ref{fig:att_vis-1} and ~\ref{fig:att_vis-2}.
{\color{black}Fig.~\ref{fig:att_vis-1} represents some visualized results of global and local attention and Fig.~\ref{fig:att_vis-2} shows the trend for attention updating every 10 epochs.}
From Fig.~\ref{fig:att_vis-1} and ~\ref{fig:att_vis-2}, we can see when predicting the related labels of an image, the model first observed the image in general (the attentive areas are covered most of the region of the image).
Then at each step of RNN, the model focused on smaller areas that may contain specific target objects.
This is very similar to human thinking that people observe an image, they always glance the whole image, and then they consider the relationships inside the image, and focus on their attention on some specific objects.

\section*{Conclusions}\label{sec:conclusion}
In this paper, we proposed a novel model that uses a global/local attention mechanism for multi-label image classification.
In our model, we first let the model focus on a more coarse area of an image, i.e., a global attention on the image.
Then, with the guidance of the global attention, the model can predict each label one by one with the local attention, which can attentively help the model focus on some specific objects.
Additionally, we proposed a joint max-margin objective that defines two max-margin in vertical and horizontal directions, respectively.
Finally, we evaluated our method on two popular multi-label image datasets, \ie, Pascal VOC 2007 and MS-COCO.
Our experimental results showed the superiority of the proposed method.

\bibliographystyle{IEEEtran}
%\bibauthoryear

\bibliography{reference}

\end{document}